\pdfoutput=1

\RequirePackage[svgnames]{xcolor}

\documentclass[11pt]{article}

\usepackage{ACL2023}

\usepackage{times}
\usepackage{latexsym}

\usepackage[T1]{fontenc}

\usepackage[utf8]{inputenc}

\usepackage{setspace}

\usepackage{microtype}

\usepackage{inconsolata}
\usepackage{graphicx}
\usepackage{caption}
\usepackage{subcaption}
\usepackage{tabularx}
\usepackage{soul}
\usepackage{float}
\usepackage{enumitem}
\usepackage{pifont}
\usepackage{arydshln}
\usepackage[ruled,linesnumbered]{algorithm2e}
\usepackage{lipsum}

\definecolor{vred}{HTML}{CE3A29}

\usepackage{tikz}

\tikzset{rndblock/.style={rounded corners,rectangle,draw,outer sep=0pt}}

\usepackage[many]{tcolorbox}

\newtcolorbox{defin}{colback=red!5!White,enhanced,title=Contributions,
	attach boxed title to top left={xshift=-4mm},boxrule=0pt,after skip=1cm,before skip=1cm,right skip=0cm,breakable,fonttitle=\bfseries,toprule=0pt,bottomrule=0pt,rightrule=0pt,leftrule=3pt,arc=0mm,skin=enhancedlast jigsaw,sharp corners,colframe=Red!55!black,colbacktitle=Red!55!black,boxed title style={
		frame code={ 
			\fill[red!25!black](frame.south west)--(frame.north west)--(frame.north east)--([xshift=3mm]frame.east)--(frame.south east)--cycle;
			\draw[line width=1mm,Red!25!black]([xshift=2mm]frame.north east)--([xshift=5mm]frame.east)--([xshift=2mm]frame.south east);
			\draw[line width=1mm,Red!25!black]([xshift=5mm]frame.north east)--([xshift=8mm]frame.east)--([xshift=5mm]frame.south east);
			\fill[Red!25!black](frame.south west)--+(4mm,-2mm)--+(4mm,2mm)--cycle;
		}
	}
}
\usetikzlibrary{shapes.geometric, arrows}
\usetikzlibrary{decorations.markings}

\definecolor{first}{RGB}{210,255,140}
\definecolor{second}{RGB}{136, 162, 190}
\definecolor{third}{RGB}{129, 222, 228}
\definecolor{fourth}{RGB}{132, 84, 246}
\definecolor{fifth}{RGB}{250, 223, 112}
\definecolor{sixth}{RGB}{203, 193, 172}
\definecolor{seventh}{RGB}{88, 112, 246}
\definecolor{eighth}{RGB}{245, 192, 106}
\definecolor{nine}{RGB}{171, 162, 111}
\definecolor{ten}{RGB}{217, 217, 217}

\definecolor{paired-light-blue}{RGB}{198, 219, 239}
\definecolor{paired-dark-blue}{RGB}{49, 130, 188}
\definecolor{paired-light-orange}{RGB}{251, 208, 162}
\definecolor{paired-dark-orange}{RGB}{230, 85, 12}
\definecolor{paired-light-green}{RGB}{199, 233, 193}
\definecolor{paired-dark-green}{RGB}{49, 163, 83}
\definecolor{paired-light-purple}{RGB}{218, 218, 235}
\definecolor{paired-dark-purple}{RGB}{117, 107, 176}
\definecolor{paired-light-gray}{RGB}{217, 217, 217}
\definecolor{paired-dark-gray}{RGB}{99, 99, 99}
\definecolor{paired-light-pink}{RGB}{222, 158, 214}
\definecolor{paired-dark-pink}{RGB}{123, 65, 115}
\definecolor{paired-light-red}{RGB}{231, 150, 156}
\definecolor{paired-dark-red}{RGB}{131, 60, 56}
\definecolor{paired-light-yellow}{RGB}{231, 204, 149}
\definecolor{paired-dark-yellow}{RGB}{141, 109, 49}
\definecolor{Teal}{RGB}{0, 50, 50}
\definecolor{White}{RGB}{250, 250, 250}

\definecolor{bg1}{HTML}{FF9966}
\definecolor{bg2}{HTML}{CCE5FF}
\definecolor{bg3}{HTML}{FFCC99}
\definecolor{bg4}{HTML}{FFC107}
\definecolor{bg5}{HTML}{FFCCCC}
\definecolor{bg6}{HTML}{D5E8D4}
\definecolor{bg7}{HTML}{eeeeee}
\definecolor{bg8}{HTML}{cdeb8b}
\definecolor{bg9}{HTML}{dae8fc}
\definecolor{bg10}{HTML}{a2e6eb}

\definecolor{bg31}{HTML}{FFCDD2} 

\definecolor{bg32}{HTML}{F8BBD0}

\definecolor{bg33}{HTML}{E1BEE7} 

\definecolor{bg34}{HTML}{D7CCC8} 

\definecolor{bg35}{HTML}{B2DFDB} 

\definecolor{bg36}{HTML}{A5D6A7} 

\definecolor{bg37}{HTML}{FFF9C4} 

\definecolor{bg38}{HTML}{FFECB3} 

\definecolor{bg111}{HTML}{CB6843}

\definecolor{bg112}{HTML}{D77C5C}

\definecolor{bg113}{HTML}{E28E6E}
\definecolor{bg114}{HTML}{E89F7D}
\definecolor{bg115}{HTML}{EDAE8A}
\definecolor{bg116}{HTML}{F0BA95}
\definecolor{bg117}{HTML}{F3C29F}
\definecolor{bg118}{HTML}{F6CCAA}
\definecolor{bg119}{HTML}{F8D5B3}
\definecolor{bg120}{HTML}{FADCBD}
\definecolor{bg121}{HTML}{FCE6C7}

\definecolor{bg39}{HTML}{FFE0B2} 

\definecolor{bg40}{HTML}{3CB371} 

\definecolor{bg43}{HTML}{ffe5d9}

\definecolor{bg15}{HTML}{7FFFD4}

\definecolor{bg17}{HTML}{F0FFFF}

\definecolor{bg18}{HTML}{F5FFFA}

\definecolor{bg19}{HTML}{F8F8FF}

\definecolor{bg20}{HTML}{FFFFFF}

\definecolor{bg21}{HTML}{E1F5FE}

\definecolor{bg22}{HTML}{B3E5FC}

\definecolor{bg23}{HTML}{81D4FA}

\definecolor{bg24}{HTML}{4FC3F7}

\definecolor{bg25}{HTML}{29B6F6}

\definecolor{bg26}{HTML}{03A9F4}

\definecolor{bg27}{HTML}{039BE5}

\definecolor{bg28}{HTML}{0288D1}

\definecolor{bg29}{HTML}{0277BD}

\definecolor{bg30}{HTML}{01579B}

\definecolor{bg16}{HTML}{FFCC99}

\definecolor{pg51}{HTML}{E8F5E9} 
\definecolor{pg52}{HTML}{C8E6C9} 
\definecolor{pg53}{HTML}{B9F6CA} 
\definecolor{pg54}{HTML}{A9DFBF} 
\definecolor{pg55}{HTML}{BCF5A6} 

\definecolor{pg56}{HTML}{BEF1CE} 
\definecolor{pg57}{HTML}{CEF6EC} 
\definecolor{pg58}{HTML}{B7F0B1} 
\definecolor{pg59}{HTML}{B1F2B5} 
\definecolor{pg60}{HTML}{9DF3C4} 

\definecolor{pg61}{HTML}{DEF7E0} 
\definecolor{pg62}{HTML}{E8F8DC} 

\definecolor{pg63}{HTML}{EBF7E7} 
\definecolor{pg64}{HTML}{F0FDF4} 

\definecolor{pg65}{HTML}{F1FEE7} 
\definecolor{pg66}{HTML}{F7FFF6} 
\definecolor{pg67}{HTML}{FCFFE7} 
\definecolor{pg68}{HTML}{F4FFD2} 
\definecolor{pg69}{HTML}{EEFFE2} 
\definecolor{pg70}{HTML}{E3FDF5} 

\definecolor{connect-color}{RGB}{0,0,0}
\definecolor{middle-color}{RGB}{255,255,255}
\definecolor{leaf-color}{RGB}{173,216,230}
\definecolor{line-color}{RGB}{25,25,112}

\usepackage{url}            
\usepackage{booktabs}       
\usepackage{amsfonts}       
\usepackage{nicefrac}       
\usepackage{microtype}      
\usepackage{xcolor}         
\usepackage{amsmath,amssymb}
\usepackage{graphicx}
\usepackage{enumitem}
\usepackage{multirow}
\usepackage{subcaption}
\usepackage{natbib}
\usepackage{listings}
\usepackage{xspace}
\usepackage{makecell}
\usepackage{soul}               
\setstcolor{red}            
\usepackage{tikz}

\usepackage{longtable}
\usepackage{tablefootnote}
\usepackage[edges]{forest}
\definecolor{hidden-draw}{RGB}{20,68,106}
\definecolor{hidden-pink}{RGB}{255,245,247}
\definecolor{red}{RGB}{255,0,0}

\definecolor{hidden-draw}{RGB}{0,0,0}
\definecolor{hidden-pink}{RGB}{255,182,193}

\usepackage{environ}

\newlength{\myl}
\expandafter\let\expandafter\origequation\csname equation*\endcsname
\expandafter\let\expandafter\endorigequation\csname endequation*\endcsname
\long\def\[#1\]{\begin{equation*}#1\end{equation*}}
\RenewEnviron{equation*}{
  \settowidth{\myl}{$\displaystyle\BODY$} 
  \origequation
    \ifdim\myl>\linewidth
      \resizebox{\linewidth}{!}{$\displaystyle\BODY$}
    \else
      \BODY 
    \fi
  \endorigequation
}

\title{\includegraphics[width=0.9\textwidth]{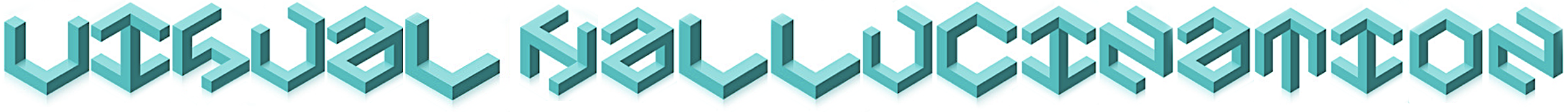}\\\textbf{}
Definition, Quantification, and Prescriptive Remediations
}
\vspace{-10mm}

\author{\textbf{Anku Rani}$^{1}$\thanks{\,\,\,Equal contribution} \quad \textbf{Vipula Rawte}$^{1}$\footnotemark[1] \quad \textbf{Harshad Sharma}$^{2}$ \quad \textbf{Neeraj Anand}$^{3}$ \quad \textbf{Krishnav Rajbangshi}$^{4}$ \\ 
\textbf{Amit Sheth}$^{1}$ \quad \textbf{Amitava Das}$^{1}$ \quad \\
$^{1}$University of South Carolina, USA \quad
$^{2}$NIT Agartala, India \quad \\ 
$^{3}$IIT Dhanbad, India \quad
$^{4}$NIT Silchar, India \quad \\
\texttt{arani@mailbox.sc.edu} \quad \texttt{amitava@mailbox.sc.edu} \quad \\
}

\let\oldtwocolumn\twocolumn
\renewcommand\twocolumn[1][]{%
    \oldtwocolumn[{#1}{
    \begin{center}
           \vspace{-6mm}
           \includegraphics[width=0.9\textwidth]{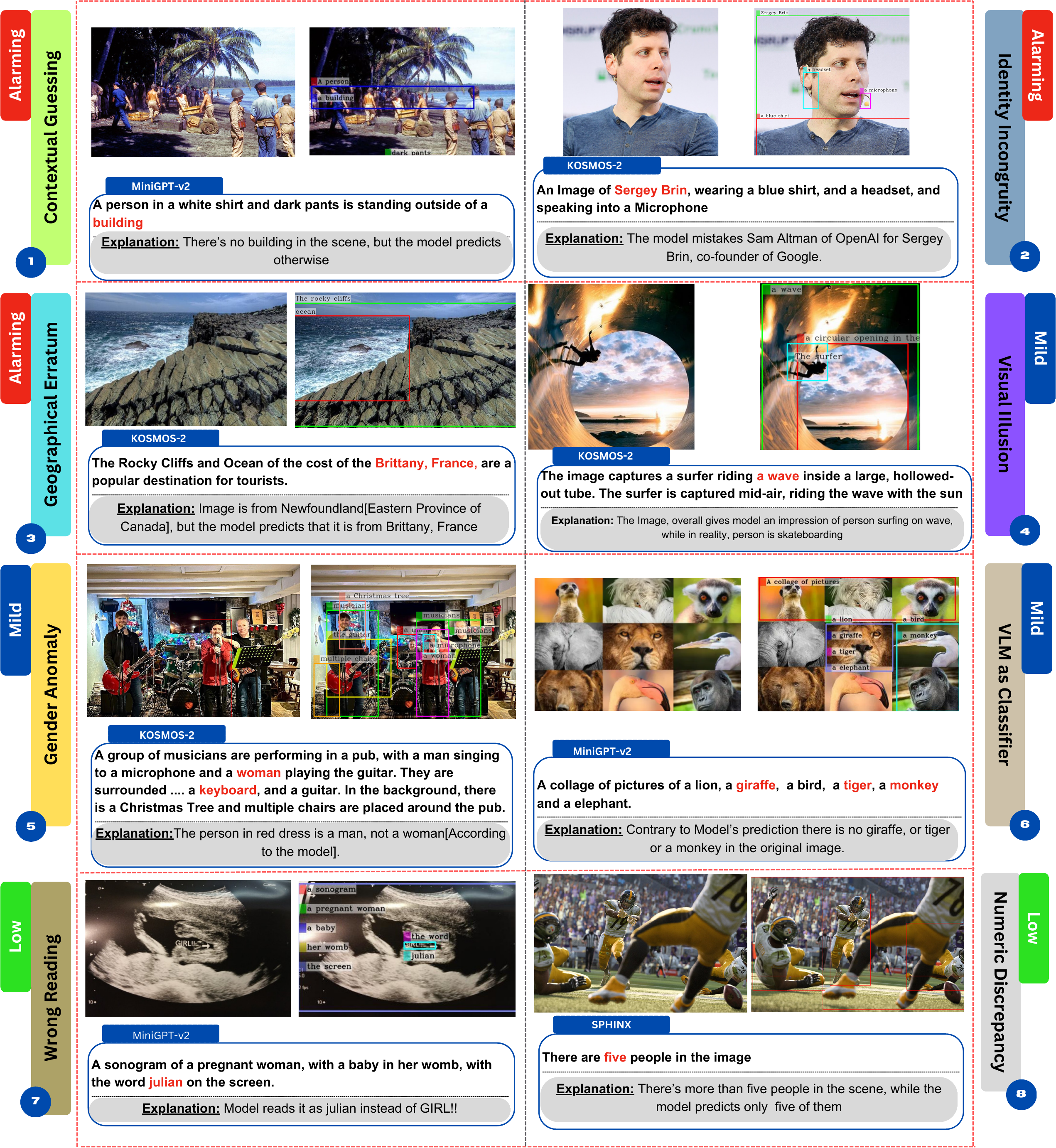}
           \vspace{-3.5mm}
           \captionof{figure}{An illustration of hallucination across your multiple categories. Here, we have used VLMs like KOSMOS-2 \cite{peng2023kosmos}, MiniGPT - v2 \cite{chen2023minigptv2}, Sphinx \cite{lin2023sphinx} to generate captions, and the text in red color represents the particular word that is hallucinating and an added line for explanation.}
           \label{fig:fig1}
        \end{center}
    }]
}

\begin{document}

\maketitle
\begin{abstract}
The troubling rise of \emph{hallucination} presents perhaps the most significant impediment to the advancement of responsible AI. In recent times, considerable research has focused on detecting and mitigating hallucination in Large Language Models (LLMs). However, it's worth noting that hallucination is also quite prevalent in Vision-Language models (VLMs). In this paper, we offer a fine-grained discourse on profiling VLM hallucination based on two tasks: i) \emph{image captioning}, and ii) \emph{Visual Question Answering (VQA)}. We delineate eight fine-grained orientations of visual hallucination: i) \emph{Contextual Guessing}, ii) \emph{Identity Incongruity}, iii) \emph{Geographical Erratum}, iv) \emph{Visual Illusion}, v) \emph{Gender Anomaly}, vi) \emph{VLM as Classifier}, vii) \emph{Wrong Reading}, and viii) \emph{Numeric Discrepancy}. We curate \textbf{V}isual \textbf{H}alluc\textbf{I}nation e\textbf{L}ici\textbf{T}ation (\includegraphics[height=0.27cm,width=1.2cm]{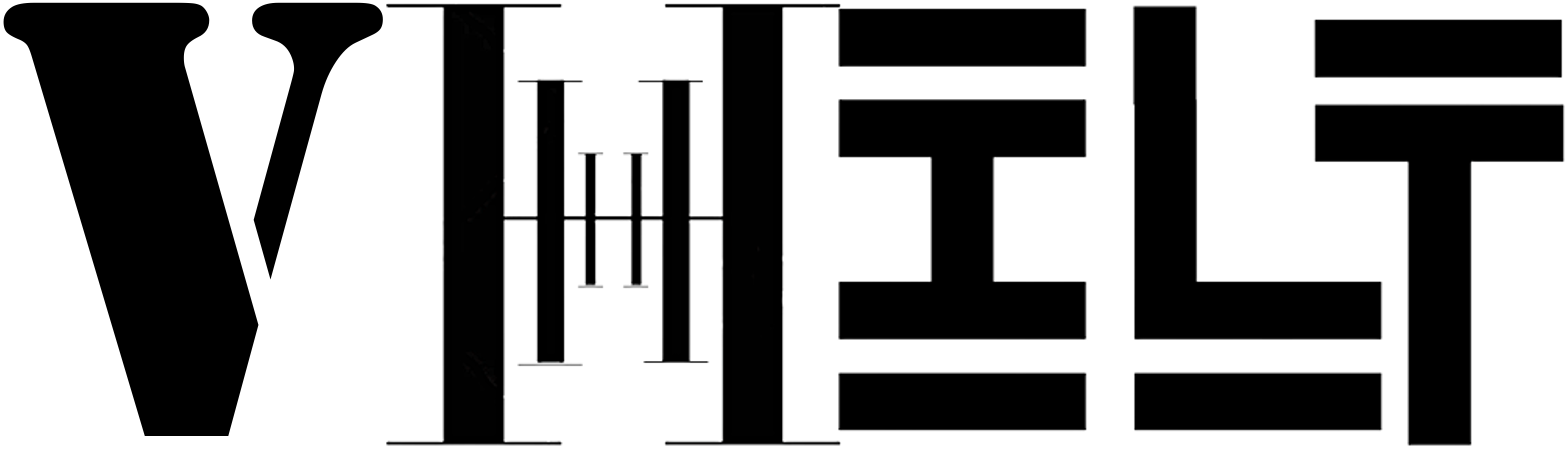}), a publicly available dataset comprising 2,000 samples generated using eight VLMs across two tasks of captioning and VQA along with human annotations for the categories as mentioned earlier.
\end{abstract}

\vspace{-10mm}
\begin{defin}

\begin{itemize}
[labelindent=-0.6em,labelsep=0.1cm,leftmargin=*]
\setlength\itemsep{0em}
\begin{spacing}{0.5}
\item[$\blacktriangleright$] 
{\footnotesize 
{\fontfamily{phv}\fontsize{8}{9}
\selectfont
Identification of Hallucination Categories: The paper identifies and categorizes various types of visual hallucinations in 8 VLMs. These include $8$ categories listed in figure \ref{fig:fig1} and section \ref{section:categorization}.}
}

\item[$\blacktriangleright$] 
{\footnotesize 
{\fontfamily{phv}\fontsize{8}{9}\selectfont
Creation of Visual Hallucination Dataset (VHiLT): The dataset comprises 2000 samples using 8 contemporary VLMs. Human annotations for the identified categories are included as well (section \ref{Section:Data})}.
}

\item[$\blacktriangleright$] 
{\footnotesize 
{\fontfamily{phv}\fontsize{8}{9}\selectfont
Mitigation Strategies: We propose three strategies to mitigate hallucinations within AI systems (section \ref{section:mitigation}).}
}
\end{spacing}
\end{itemize}
\end{defin}
\vspace{-9mm}

\section{Visual Hallucination - an extensive categorization} \label{section:categorization}
Despite the rapid advances in Generative AI, policymakers \cite{Janjeva_Harris_Sarah_Kasprzyk_Gausen} are primarily concerned with the issue of hallucinations. These occurrences of \emph{hallucinations} pose a significant risk of eroding trust in technology. For instance, when Google's Bard AI "hallucinated" during its initial public demonstration, Alphabet experienced a temporary loss of \$100 billion in market value \cite{Olson_2023}. 

The study of hallucinations for LLMs has recently attracted considerable attention \cite{rawte-etal-2023-troubling, tonmoy2024comprehensive}. This paper delves into visual hallucination, a phenomenon notably prevalent in numerous recent VLMs. Given that this field is still emerging, it is imperative to initially comprehend, classify, and quantify these phenomena while establishing a benchmark. This will aid the scientific community in collectively addressing this issue. Therefore, this paper aims to provide a comprehensive categorization of VLM hallucinations. We defined eight categories of Visual Hallucination: 

\vspace{-4mm}

\begin{itemize}

\item[\ding{224}] \begin{tcolorbox}[enhanced,attach boxed title to top right={yshift=-2.5mm,yshifttext=-1mm}, left=1pt,right=1pt,top=1pt,bottom=1pt,width=40mm,height=8mm,
  colback=first!,colframe=first!25!first,colbacktitle=red!80!black,
  title=alarming,fonttitle=\ttfamily\bfseries\scshape\fontsize{9}{9.6}\selectfont,
  boxed title style={size=fbox,colframe=red!50!black} ]
  {\textbf{\footnotesize Contextual Guessing (CG)}}
\end{tcolorbox} 
\vspace{-7mm}

\hspace{42mm}{\small When the model generates unrelated elements that bear no resemblance to the subject at hand, highlighting the non-deterministic nature of the model.}

\vspace{-3mm}

\item[\ding{224}] \begin{tcolorbox}[enhanced,attach boxed title to top right={yshift=-2.5mm,yshifttext=-1mm}, left=1pt,right=1pt,top=1pt,bottom=1pt,width=40mm,height=8mm,
  colback=second!,colframe=second!25!second,colbacktitle=red!80!black,
  title=alarming,fonttitle=\ttfamily\bfseries\scshape\fontsize{9}{9.6}\selectfont,
  boxed title style={size=fbox,colframe=red!50!black} ]
  {\textbf{\footnotesize Identity Incongruity (II)}}
\end{tcolorbox} 
\vspace{-7mm}
\hspace{42mm}{\small It's when the model can't differentiate between a person's real and fake identity traits, causing a mismatch with the predicted identity.}
\vspace{-7mm}

\item[\ding{224}] \begin{tcolorbox}[enhanced,attach boxed title to top right={yshift=-2.5mm,yshifttext=-1mm}, left=1pt,right=1pt,top=1pt,bottom=1pt,width=40mm,height=8mm,
  colback=third!,colframe=third!25!third,colbacktitle=red!80!black,
  title=alarming,fonttitle=\ttfamily\bfseries\scshape\fontsize{9}{9.6}\selectfont,
  boxed title style={size=fbox,colframe=red!50!black} ]
  {\textbf{\footnotesize Geographic Erratum  (GE)}}
\end{tcolorbox} 
\vspace{-7mm}
\hspace{42mm}{\small  In this scenario, the model produces an inaccurate prediction or guess related to the geographical location or landmark of the place under consideration.}
\vspace{-3mm}

\item[\ding{224}] \begin{tcolorbox}[enhanced,attach boxed title to top right={yshift=-2.5mm,yshifttext=-1mm}, left=1pt,right=1pt,top=1pt,bottom=1pt,width=40mm,height=8mm,
  colback=fourth!,colframe=fourth!25!fourth,colbacktitle=blue!80!black,
  title=mild,fonttitle=\ttfamily\bfseries\scshape\fontsize{9}{9.6}\selectfont,
  boxed title style={size=fbox,colframe=red!50!black} ]
  {\textbf{\footnotesize Visual Illusion  (VI)}}
\end{tcolorbox} 
\vspace{-5mm}
\hspace{42mm}{\small   The model can be misled, creating a distorted perception that deviates from reality causing the model's output to be partially inaccurate due to a specific aspect of the image.}
\vspace{-4mm}

\item[\ding{224}] \begin{tcolorbox}[enhanced,attach boxed title to top right={yshift=-2.5mm,yshifttext=-1mm}, left=1pt,right=1pt,top=1pt,bottom=1pt,width=40mm,height=8mm,
  colback=fifth!,colframe=fifth!25!fifth,colbacktitle=blue!80!black,
  title=mild,fonttitle=\ttfamily\bfseries\scshape\fontsize{9}{9.6}\selectfont,
  boxed title style={size=fbox,colframe=red!50!black} ]
  {\textbf{\footnotesize Gender Anomaly  (GA)}}
\end{tcolorbox}
\vspace{-7mm}
\hspace{42mm} {\small The model provides an inaccurate representation of gender identity.}
\vspace{-2mm}

\item[\ding{224}] \begin{tcolorbox}[enhanced,attach boxed title to top right={yshift=-2.5mm,yshifttext=-1mm}, left=1pt,right=1pt,top=1pt,bottom=1pt,width=40mm,height=8mm,
  colback=eighth!,colframe=eighth!25!eighth,colbacktitle=blue!80!black,
  title=mild,fonttitle=\ttfamily\bfseries\scshape\fontsize{9}{9.6}\selectfont,
  boxed title style={size=fbox,colframe=blue!50!black} ]
  {\textbf{\footnotesize VLM as Classifier  (VC)}}
\end{tcolorbox} 
\vspace{-7mm}
\hspace{40mm} {\small This is a situation where the model's proficiency is assessed based on its ability to differentiate between two/more entities.}
\vspace{-2mm}

\item[\ding{224}] \begin{tcolorbox}[enhanced,attach boxed title to top right={yshift=-2.5mm,yshifttext=-1mm}, left=1pt,right=1pt,top=1pt,bottom=1pt,width=40mm,
  colback=nine!,colframe=nine!25!nine,colbacktitle=green!80!black,
  title=low,fonttitle=\ttfamily\bfseries\scshape\fontsize{9}{9.6}\selectfont,
  boxed title style={size=fbox,colframe=red!50!black} ]
  {\textbf{\footnotesize Wrong Reading  (WR)}}
\end{tcolorbox} 
\vspace{-7mm}
\hspace{41mm} {\small When a text is engraved in an image, but the VLM read it wrong.}
\vspace{-2mm}

\item[\ding{224}] \begin{tcolorbox}[enhanced,attach boxed title to top right={yshift=-2.5mm,yshifttext=-1mm}, left=1pt,right=1pt,top=1pt,bottom=1pt,width=40mm,
  colback=ten!,colframe=ten!25!ten,colbacktitle=green!80!black,
  title=low,fonttitle=\ttfamily\bfseries\scshape\fontsize{9}{9.6}\selectfont,
  boxed title style={size=fbox,colframe=red!50!black} ]
  {\textbf{\footnotesize Numeric Discrepancy  (ND)}}
\end{tcolorbox} 
\vspace{-7mm}
\hspace{41mm} {\small When the model encounters difficulty accurately counting the number of entities within the analyzed image leading to an inaccurate count.}

\end{itemize}
\vspace{-2mm}

VLMs are predominantly employed for two purposes: i) image captioning, and ii) Visual Question Answering. Hence, in curating the \includegraphics[height=0.27cm,width=1.2cm]{img/VHILT.png} dataset, we have incorporated both of these applied aspects.

\subsection{Caption Hallucination} \label{subsection:Caption Hallucination}
Caption Hallucination, within the realm of VLMs, occurs when the VLM, after being provided with an image as input, produces descriptions or captions that either contain elements inconsistent with the image or omit important aspects of it as discussed in fig \ref{fig:fig1}. This is otherwise known as \emph{object hallucination}. The reasons behind this kind of hallucination, are multifaceted. Recent studies \cite{biten2022let,li2023evaluating,zhou2023analyzing} have demonstrated that factors such as \emph{co-occurrence} and \emph{uncertainty} are significant contributors. Additionally, the lack of alignment between visual and language annotations \cite{zhai2023halle}, insufficient training objectives \cite{chen2023mitigating}, and language bias \cite{guan2023hallusionbench} have been identified as prominent factors. It is evident that researchers have yet to make a consensus on the causality of visual hallucination, but prevalence of visual hallucination in VLMs necessitates to study of the phenomena further. 
 

\subsection{VQA Hallucination} \label{subsection:VQA Hallucination}
\begin{figure}[!ht]
    \centering
\includegraphics[width=0.5\textwidth]{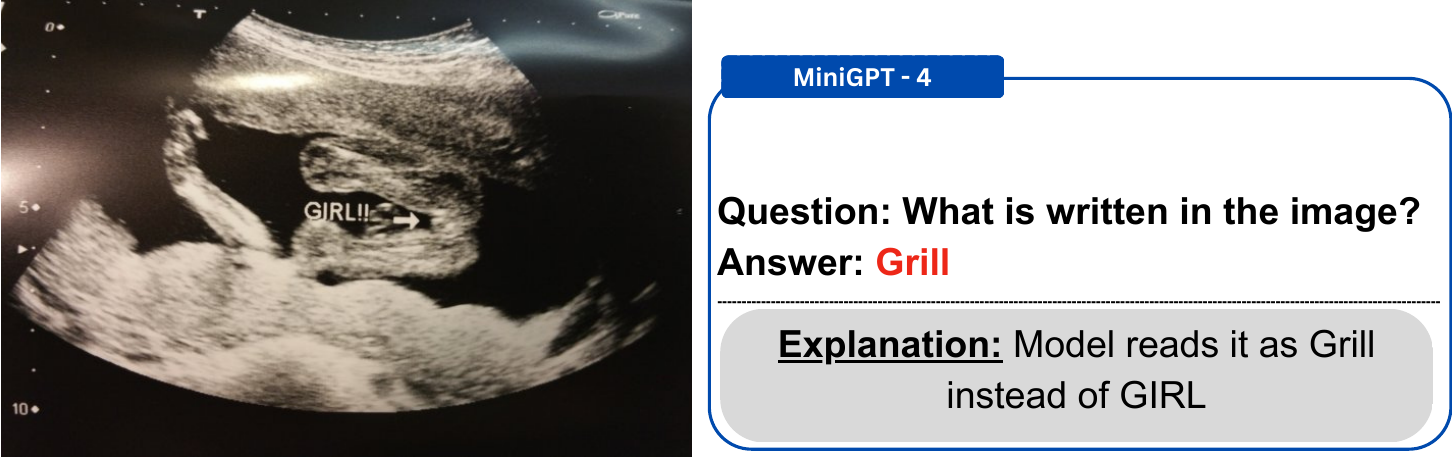}
\caption{This describes the case of hallucination in the VQA task. The picture is processed by MiniGPT-4, and the model is asked about the content of the image. The response provided is "GRILL," falling into the hallucination category of Wrong Reading.}
    \label{fig:VQA_Img1}
\end{figure}

\begin{figure}[h!]
     \centering
     \begin{subfigure}[b]{0.45\textwidth}
         \centering
         \includegraphics[width=\textwidth]{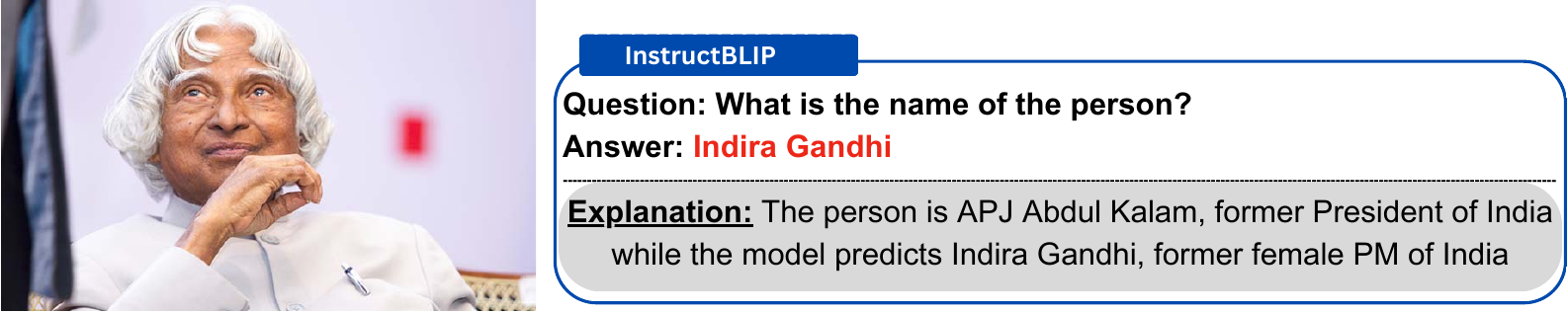}
         \caption{Identity Incongruity}
     \end{subfigure}
     \hfill
     \begin{subfigure}[b]{0.45\textwidth}
         \centering
         \includegraphics[width=\textwidth]{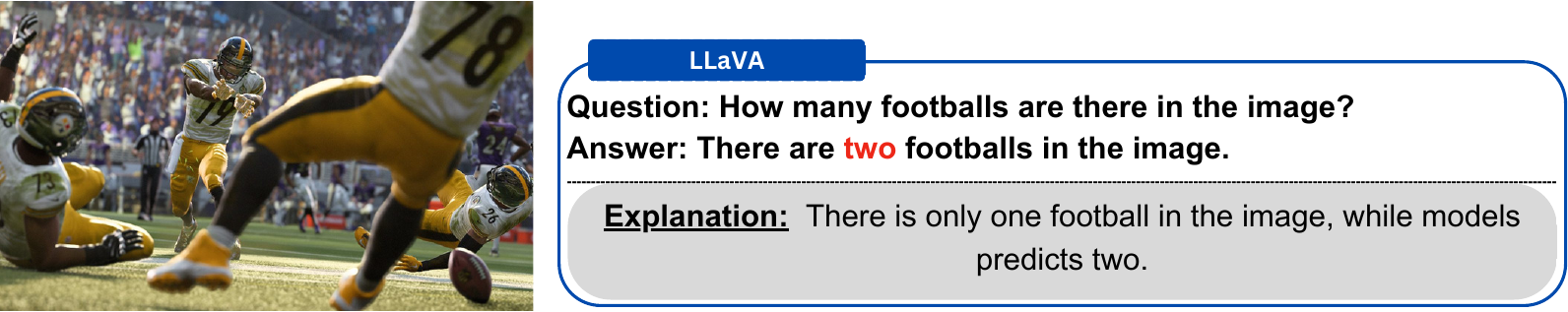}
         \caption{Numeric Discrepancy}
     \end{subfigure}
     \hfill
     \begin{subfigure}[b]{0.45\textwidth}
         \centering
         \includegraphics[width=\textwidth]{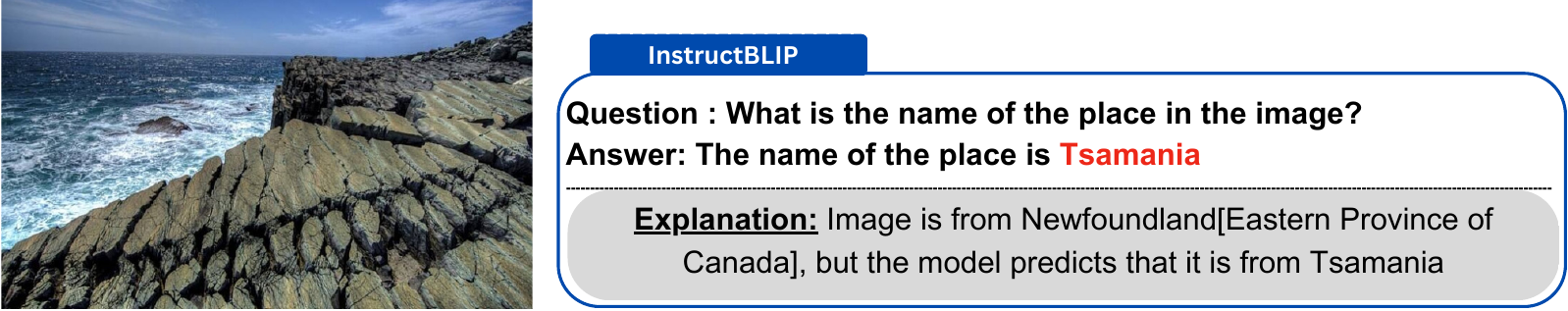}
         \caption{Geographic Erratum}
     \end{subfigure}
     \vspace{-2mm}
        \caption{In the VQA task, we display examples across three categories, showcasing instances where the model produces hallucinatory outputs. Explanations for these are provided in each figure, with additional examples detailed in Appendix section \ref{App:exampleset from VQA}.}
        \label{fig:vqa_hallucination}
\end{figure}

In the domain of Visual Question Answering (VQA) for VLMs, hallucinations may appear as responses that contain references or descriptions that are incorrect \cite{gunjal2023detecting}. Unlike captioning, VQA requires both understanding the visual content of an image and comprehending the questions posed about it as presented in figure \ref{fig:VQA_Img1} and figure \ref{fig:vqa_hallucination}. Hallucinations in VQA may arise from the model's inability to accurately interpret visual cues, grasp the context of the question, or generate suitable responses based on image features. Moreover, VQA systems may encounter challenges in complex object reasoning, where they struggle to correctly deduce intricate relationships between objects or scenes in the image, leading to inaccurate answers.

\vspace{-5mm}
\section{\includegraphics[height=0.27cm,width=1.2cm]{img/VHILT.png} dataset} \label{Section:Data}

The proliferation of Generative AI models has led to a significant surge in online misinformation, as outlined in the recent report \cite{EU_code_of_conduct_2022} from the EU. Building on this inspiration, we aim to focus our collection of visual hallucination data on the news domain. As we annotate the hallucinations generated by VLMs, we also require gold standard, factually correct captions or answers. Hence, we opt for the \emph{New York Times Twitter handle} \cite{nytimes_twitter} as the trusted source of our factual data, over a period ranging from 2011 to 2021, covering a widespan multimodal data of 10 years. NYT tweets, being authored by professional journalists, are grammatically correct, devoid of wordplay, incomplete sentences, or other typical issues found in Twitter data.

We exclusively selected tweets containing images to investigate visual hallucination. Subsequently, we implemented a thorough data filtration process to refine the dataset. This process entailed eliminating duplicate tweets and irrelevant content, such as tweets related to word games (e.g., Wordles), crossword puzzles, spelling bees, etc. This ensured that only original and relevant information was retained for analysis. Furthermore, the filtration process involved removing tweets with non-English text. Furthermore, we removed hashtags and URLs, retaining only alphanumeric characters. We chose a random sample of 2000 datapoints to study visual hallucination.

\subsection{Choice of VLMs: Rationale and Coverage} \label{subsection:choice}
When selecting VLMs, we meticulously opted for state-of-the-art (SoTA) models for both tasks: i) image captioning, and ii) Visual Question Answering. The \ul{image captioning-based models} comprise: (i) Kosmos-2 \cite{peng2023kosmos}, (ii) MiniGPT-V2 \cite{chen2023minigptv2}, and (iii) Sphinx \cite{lin2023sphinx}. The \ul{VQA-based models} include: (i) LLaVa \cite{liu2023visual}, (ii) MiniGPT-4 \cite{zhu2023minigpt}, (iii) InstructBLIP \cite{dai2305instructblip}, (iv) MultimodalGPT \cite{gong2023multimodal}, and (v) MplugOwl \cite{ye2023mplug}. Appendix \ref{App:choice} discusses additional details about our selection criteria. Given the ever-evolving nature of the field, and \includegraphics[height=0.27cm,width=1.2cm]{img/VHILT.png} benchmark leaderboards will remain accessible to the research community, fostering an environment of continuous updates and contributions.

\noindent
\textbf{Caption hallucination}: 
We began by using NYT news images and fed them into Kosmos-2, MiniGPT-V2, and Sphinx to generate text captions. At this point, we have the image, caption generated by VLMs, and the actual tweet aka news headline associated with the image obtained from NYT. We also have bounding boxes and grounding information obtained from the VLMs. We provided all this information to our in-house annotators and asked them two questions: i) \emph{Do you observe any visual hallucinations in this VLM-generated caption? Please annotate it at the sentence level}. It's worth noting that text captions may contain multiple sentences. ii) \emph{If there is a visual hallucination, could you please describe its type?} Four in-house annotators were involved in the annotation process. After annotating 2000 instances, they collectively discussed and finalized the eight categories.

\noindent
\textbf{VQA hallucination}: After finalizing all eight categories, we presented NYT images to the annotators and instructed them to compose specific questions that could potentially lead to visual hallucinations corresponding to those categories. For example, We recorded responses to those specific questions from all five VQA models: i) LLaVa, ii) MiniGPT-4, iii) InstructBLIP, iv) MultimodalGPT, and v) MplugOwl. Then, we asked annotators to indicate whether they detected any visual hallucination in the answers generated by those VLMs.

In summary, we observed two key points: i) There are instances where two or more hallucination categories are present, leading to confusion among annotators. We deliberately avoided multi-class classification at this point, ii) Additionally, we identified new types of hallucination beyond the eight prevalent categories. We deliberately excluded such instances with skewed categorical examples, as we believe they are rare cases and our focus is on investigating prevalent visual hallucination categories.

\subsection{Annotation Process} \label{subsection:data annotation}
The annotation process involved two steps. Initially, we annotated a small amount of data in-house, and then we expanded this process using the crowdsourcing service Amazon Mechanical Turk (AMT).

\noindent
\textbf{Pilot in-house annotation}
We conducted an extensive in-house study to categorize visual hallucinations, annotating 2,000 datapoints for both image captioning and VQA tasks. 

\noindent
\textbf{Web interface for annotation}: To facilitate the annotation process for the annotators, it is crucial to provide them with a user-friendly interface that enables easy navigation. Figure \ref{fig:webinterface_Caption} and \ref{fig:webinterface_vqa} (c.f Appendix \ref{App:DataAnnotation webinterface}) shows our annotation web interface used to construct the VHILT dataset for studying hallucination in captioning and VQA technique respectively. 


\noindent
\textbf{Selecting quality annotators on AMT}: It is widely acknowledged that platforms like AMT can be noisy, making the selection of high-quality annotators a critical step in ensuring accurate annotations. The in-house annotation of 2,000 data points played a significant role in achieving this goal. To identify reliable annotators, we initiated a pilot task and only selected those with an accuracy rate of over 90\% based on our in-house annotated dataset. Further discussion regarding compensation etc please refer to Appendix section \ref{App:Data Annotation}.





\section{Hallucination Mitigation - an emerging research paradigm} \label{section:mitigation}

\tikzset{
  my-box/.style={
    rectangle,
    draw=hidden-draw,
    rounded corners,
    text opacity=1,
    minimum height=1.5em,
    minimum width=40em,
    inner sep=2pt,
    align=center,
    line width=0.8pt,
  },
  leaf/.style={
    my-box,
    minimum height=1.5em,
    text=black,
    align=center,
    font=\normalsize,
    inner xsep=2pt,
    inner ysep=4pt,
    line width=0.8pt,
  }
}

\begin{figure*}[!htb]
  \centering
  \resizebox{\textwidth}{!}{%
    \begin{forest}
      forked edges,
      for tree={
        grow=east,
        reversed=true,
        anchor=base west,
        parent anchor=east,
        child anchor=west,
        base=center,
        font=\large,
        rectangle,
        draw=hidden-draw,
        rounded corners,
        align=center,
        text centered,
        minimum width=5em,
        edge+={darkgray, line width=1pt},
        s sep=3pt,
        inner xsep=2pt,
        inner ysep=3pt,
        line width=0.8pt,
        ver/.style={rotate=90, child anchor=north, parent anchor=south, anchor=center},
      },
      where level=1{text width=15em,font=\normalsize,}{},
      where level=2{text width=14em,font=\normalsize,}{},
      where level=3{minimum width=10em,font=\normalsize,}{},
      where level=4{text width=26em,font=\normalsize,}{},
      where level=5{text width=20em,font=\normalsize,}{},
      [
        \textbf{Hallucination Mitigation}\\ \textbf{Techniques in VLMs}, for tree={fill=paired-light-red!70}
        [
          \textbf{Data-Driven Approaches}, for tree={fill=paired-light-yellow!45}
            [
            \textbf{M-Hal Detect} \cite{gunjal2023detecting} \\
            \textbf{LRV-Instruction} \cite{liu2023aligning}, text width=60em, for tree={fill=bg1}, leaf
            ]
        ]
        [
          \textbf{Training Adjustments}, for tree={fill=bg22}
          [
            \textbf{Introducing New Decoding} \\ \textbf{Strategy}, for tree={fill=bg39},text width=14em
            [
              \textbf{VCD} \cite{leng2023mitigating} \\, text width=44.3em, for tree={fill=pg67}, leaf
            ]
          ]
          [
            \textbf{Changing the Pre-Training} \\ \textbf{Objective},text width=14em, for tree={fill=bg39}
            [
              \textbf{ObjMLM} \cite{dai2022plausible}, text width=44.3em, for tree={fill=pg60}, leaf
            ]
          ]
          [
            \textbf{Visual Grounding} \\ \textbf{Integration}, text width=14em, for tree={fill=bg39}
            [
              \textbf{MoF} \cite{tong2024eyes}, text width=44.3em, for tree={fill=bg35}, leaf
            ]
          ]
        ]
        [
          \textbf{Post-Processing Techniques}, for tree={fill=pg58}
            [
            \textbf{Woodpecker} \cite{yin2023woodpecker} \\
            \textbf{Volcano} \cite{lee2023volcano} \\
            \textbf{LURE} \cite{zhou2023analyzing}, text width=60em, for tree={fill=bg33}, leaf
            ]
        ]
      ]
    \end{forest}
  }
  \vspace{-5mm}
  \caption{Taxonomy of hallucination mitigation techniques in VLMs, showcasing various data-driven, training adjustments and post-processing techniques for mitigating hallucinations in VLMs.} 
  \label{fig:mitigation _strategies}
\end{figure*}

Visual hallucination has recently garnered research attention. However, to our knowledge, none of the previous studies have delved into a comprehensive categorization of visual hallucinations like ours. We identify three major families of mitigation techniques proposed so far: i) \emph{data-driven approaches}, ii) \emph{training adjustments}, and iii) \emph{post-processing techniques}. A detailed taxonomy of this classification is presented in Figure ~\ref{fig:mitigation _strategies}.

\noindent
\textbf{Data-driven approaches}: 
The data-driven approach to mitigate hallucination involves acquiring manually labeled visual hallucination data, followed by fine-tuning or devising techniques based on this data to mitigate hallucination in VLMs. The techniques proposed here partially fall under data-driven techniques, as we have also created the \includegraphics[height=0.27cm,width=1.2cm]{img/VHILT.png} dataset.
\cite{gunjal2023detecting} presented M-HalDetect, a multi-modal hallucination detection dataset emphasizing detailed image descriptions. The authors utilized images from popular datasets like MS COCO and employed VQA models such as InstructBLIP to generate detailed descriptions of given images with the instruction "\emph{Provide an intricate description of the image, capturing its visual elements, including colors}". The generated descriptions were then manually examined to identify any hallucinations. As mitigation techniques, the authors proposed two reinforcement learning-based reward models: a) a multimodal reward model, and b) Fine-grained Direct Preference Optimization (FDPO). They fine-tuned a VQA model based on these reward models and claimed to achieve approximately 50\% reduction in hallucination. This study closely relates to the topic discussed in our paper; however, our study offers a more comprehensive categorization of hallucination. We examined three captioning models and five different VQA models. Additionally, we argue that fine-tuning reward models on similar data may lead to overfitting. Therefore, whether this 50\% improvement is transferrable to other domains, such as news in our case, remains an open question.

\noindent
In \cite{liu2023aligning}, the authors introduced the LRV-Instruction tuning dataset, consisting of 120k visual instructions generated by GPT4 across 16 vision-and-language tasks. They categorized visual hallucination into two classes: Nonexistent Element Manipulation and Existent Element Manipulation. Their proposed method, GPT4-Assisted Visual Instruction Evaluation (GAVIE), aims to evaluate visual instruction tuning without human-annotated ground truth. While promising, we advocate for a more detailed categorization of visual hallucination. Additionally, our manual crafting of categorical questions offers advantages over GPT4-based automatic question creation.


\noindent
\textbf{Training adjustments}: Training adjustments encompass modifications made to the training process of VLMs with the goal of enhancing their ability to generate fewer hallucinations. There are three kinds of major training adjustments proposed so far - a) new decoding strategy, b) changing pre-training objective and c) visual grounding integration. 
In \cite{leng2023mitigating}, Visual Contrastive Decoding (VCD) was introduced to address object hallucinations by introducing controlled uncertainty using Gaussian noise. VCD involves three stages: injecting Gaussian noise into the original image, generating two sets of textual outputs—one for the original image and one for the distorted version using dual-stream decoding, and calculating a contrastive distribution to sample words consistent with both versions of the image, thereby reducing object hallucinations. However, it's important to note that this approach specifically targeted object hallucinations.
In \cite{dai2022plausible}, ObjMLM was introduced as a straightforward yet effective approach to enhance object-level image-text alignment and mitigate object hallucination. ObjMLM functions similarly to Masked Language Modeling (MLM) loss, but it specifically targets objects mentioned in the text that also appear in the input image. By masking out these objects during training, ObjMLM guides the model to generate only objects that correspond to those in the image, thereby reducing the occurrence of false or nonexistent objects. It's worth mentioning that this method focuses exclusively on object hallucination.
In \cite{tong2024eyes}, Mixture-of-Features (MoF) was proposed to address hallucination and enhance visual grounding in Multimodal LLMs. This method integrates features from different visual encoders, such as the pretrained Contrastive Language-Image Pre-Training (CLIP) model \cite{radford2021learning}, known for its text-image alignment, and the vision-only self-supervised DINOv2 model \cite{oquab2023dinov2}, renowned for fine-grained recognition. They introduced two methods of integration: Additive-MoF (A-MoF), which linearly combines features from CLIP and DINOv2, prioritizing DINOv2 for better grounding but sacrificing instruction-following ability. Interleaved-MoF (I-MoF) spatially mixes visual tokens from both CLIP and DINOv2 models, achieving robust grounding while preserving instruction-following capabilities. While this paper does touch upon visual hallucination to some extent, it's important to note that the aforementioned techniques are solely evaluated on visual grounding.

\noindent
\textbf{Post-processing techniques}: Post-processing techniques involve methods applied after the model's inference process to refine and correct its outputs. This method family can be compared with Retrieval Augmented Generation (RAG), which is quite popular for hallucination mitigation in LLMs.

In \cite{yin2023woodpecker}, Woodpecker was introduced as a training-free method employing a five-stage framework for diagnosing and refining model responses to correct hallucinations. The primary aim of Woodpecker is to enhance the factuality of the generated content from MLLMs by refining the responses and modifying the hallucinated parts.
In \cite{lee2023volcano}, VOLCANO was introduced as a technique aimed at mitigating multimodal hallucination by leveraging self-feedback. VOLCANO operates by iteratively refining model responses using both natural language feedback and visual information. Initially, the method generates a response based on the provided image and question. Subsequently, it iteratively revises this response until further improvement is deemed unnecessary. This iterative refinement process is guided by self-feedback, enabling the model to correct hallucinated responses by incorporating detailed visual information. 
In \cite{zhou2023analyzing}, LVLM Hallucination Revisor (LURE) was proposed as a post-processing algorithm aimed at rectifying object hallucination in LVLMs. LURE is designed to rectify object hallucination by revising the generated descriptions to align more closely with the actual content of the images. To create the revisor, the authors first generate a hallucinatory dataset using GPT-3.5 by making two modifications to the original correct captions. First, they insert additional object texts into the description that are likely to co-occur with the objects contained in the initial description. This modification allows LURE to learn to disentangle such co-occurrence patterns effectively. Second, they replace uncertain objects or those at the end of descriptions with a placeholder tag, encouraging the revisor to re-evaluate these objects. Once trained on the hallucinatory dataset, the revisor can seamlessly integrate with any LVLM to correct potential hallucinatory descriptions.

\vspace{-2mm}
\section{Conclusion}
The enthusiasm and achievements surrounding Generative AI models have led to their widespread adoption, and this trend is only expected to flourish. However, one of the most significant challenges faced by these models today is hallucination. 
\includegraphics[height=0.27cm,width=1.2cm]{img/VHILT.png} benchmark will be publicly open for further collaborative updates.

\section{Ethics Statement}
Through our experiments, we have uncovered the susceptibility of VLMs to hallucination. 
However, we must address the potential misuse of our findings by malicious entities who may exploit AI-generated images, such as creating indistinguishable fake news from human-written content. We vehemently discourage such misuse and strongly advise against it.

\bibliography{custom}
\bibliographystyle{acl_natbib}
\newpage
\onecolumn
\appendix
\setcounter{section}{0}

\section*{Appendix}\label{sec:appendix}
This section provides additional examples to assist in the understanding and interpretation of the research work presented.

\section{Details on Choice of VLMs: Rationale and Coverage} \label{App:choice}

We shortlisted five SOTA models for VQA InstructBlip\cite{dai2305instructblip}, MiniGPT - v2\cite{chen2023minigptv2}, Multimodal-gpt\cite{gong2023multimodal}, LLava\cite{liu2023visual}, mPlug-Owl\cite{ye2023mplug}. Recent work on visual hallucination in VLMs chooses these models LURE\cite{zhou2023analyzing}, POPE\cite{li2023evaluating}, and HaELM\cite{wang2023evaluation} for analysis. In a similar line of reasoning for the captioning task, we shortlisted three SOTA models for studying hallucination in captioning, namely Kosmos-2\cite{peng2023kosmos}, MiniGPT-v2\cite{chen2023minigptv2}, and SPHINX\cite{lin2023sphinx}.

\subsection{Model Hyperparameter Details}

In table \ref{tab:hyperparameter_captioning}, we cover hyperparameter details of Kosmos-2, MiniGPT-v2, and Sphinx for the captioning task.

\begin{table*}[!htp]
\centering
\resizebox{0.45\textwidth}{!}{%
\begin{tabular}{c|cc}
\hline
Model Name                  & \multicolumn{2}{c}{Hyperparameters}     \\ \hline
\multirow{4}{*}{Kosmos-2}   & Image Embedding Number          & 64         \\
                            & Learning rate                   & 2e-4       \\
                            & Adam                            & (0.9,0.98) \\
                            & weight decay                    & 0.01       \\ \hline
\multirow{2}{*}{MiniGPT-v2} & Temperature                     & 0.6        \\
                            & Learning rate                   & 1e-4       \\ \hline
\multirow{4}{*}{Sphinx}     & Single Turn Max Response Length & 1024       \\
                            & Temperature                     & 0.1        \\
                            & Top-p                           & 0.75       \\
                            & Adam                            & (0.9,0.95) \\ \hline
\end{tabular}%
}
\caption{Hyperparameters for different models for captioning task}
\label{tab:hyperparameter_captioning}
\end{table*}

\vspace{-4mm}
In table \ref{tab:hyperparameter_vqa}, we cover hyperparameter details of InstructBlip, MiniGPT - v2, Multimodal-gpt, LLava, and mPlug-Owl for the VQA task.

\begin{table*}[!htp]
\centering
\resizebox{0.4\textwidth}{!}{%
\begin{tabular}{c|cc}
\hline
Models                          & \multicolumn{2}{c}{Hyperparameters} \\ \hline
\multirow{9}{*}{InstructBLIP}   & Number of Beams           & 5            \\
                                & Max Length                & 256          \\
                                & minimum length            & 1            \\
                                & top p                     & 0.9          \\
                                & repetition penalty        & 1.5          \\
                                & length penalty            & 1            \\
                                & Temperature               & 1            \\
                                & Adam                      & (0.9,0.999)  \\
                                & weight decay              & 0.05         \\ \hline
\multirow{11}{*}{mPlug-Owl}     & max length                & 500          \\
                                & Temperature               & 0.75         \\
                                & top\_p                     & 1            \\
                                & top\_k                     & 25           \\
                                & penalty\_alpha             & 0.25         \\
                                & repetition penalty        & 1            \\
                                & Number\_repeat\_ngram\_size  & 0            \\
                                & seed                      & -1           \\
                                & Adam                      & (0.9,0.98)   \\
                                & learning rate             & 0.0001       \\
                                & weight decay              & 0.1          \\ \hline
\multirow{2}{*}{MiniGPT-4}      & Beam Search Number        & 1            \\
                                & Temperature               & 1            \\ \hline
\multirow{3}{*}{LLaVa}          & Maximum context length    & 2048         \\
                                & learning rate             & 2e-5         \\
                                & weight decay              & 0            \\ \hline
\multirow{6}{*}{Multimodal-gpt} & maximum new token         & 200          \\
                                & num\_beams                 & 1            \\
                                & temperature               & 1            \\
                                & top\_k                     & 0            \\
                                & top\_p                     & 0            \\
                                & learning rate             & 1e-5         \\ \hline
\end{tabular}
}
\caption{Hyperparameters for different models for VQA task}
\label{tab:hyperparameter_vqa}
\end{table*}

\newpage

\section{Data Annotation} \label{App:Data Annotation}

Here we cover additional details about the Data Annotation step expanding further on top of section \ref{subsection:data annotation} where we mentioned the use of AMT and the web interface. Another crucial consideration when annotating data on crowdsourcing platforms is the compensation offered to annotators. While offering too little may deter interest, excessively high wages may attract undesirable spammers. Achieving the ideal balance required several rounds of iteration to determine an appropriate compensation scheme. By carefully addressing factors such as selecting qualified annotators and establishing suitable compensation rates, we improved the quality of annotations obtained from crowdsourcing platforms.

\subsection{Web Interface for Annotation for Captioning task}\label{App:DataAnnotation webinterface}

\begin{figure*}[!ht]
    \centering
\includegraphics[width=0.8\textwidth]{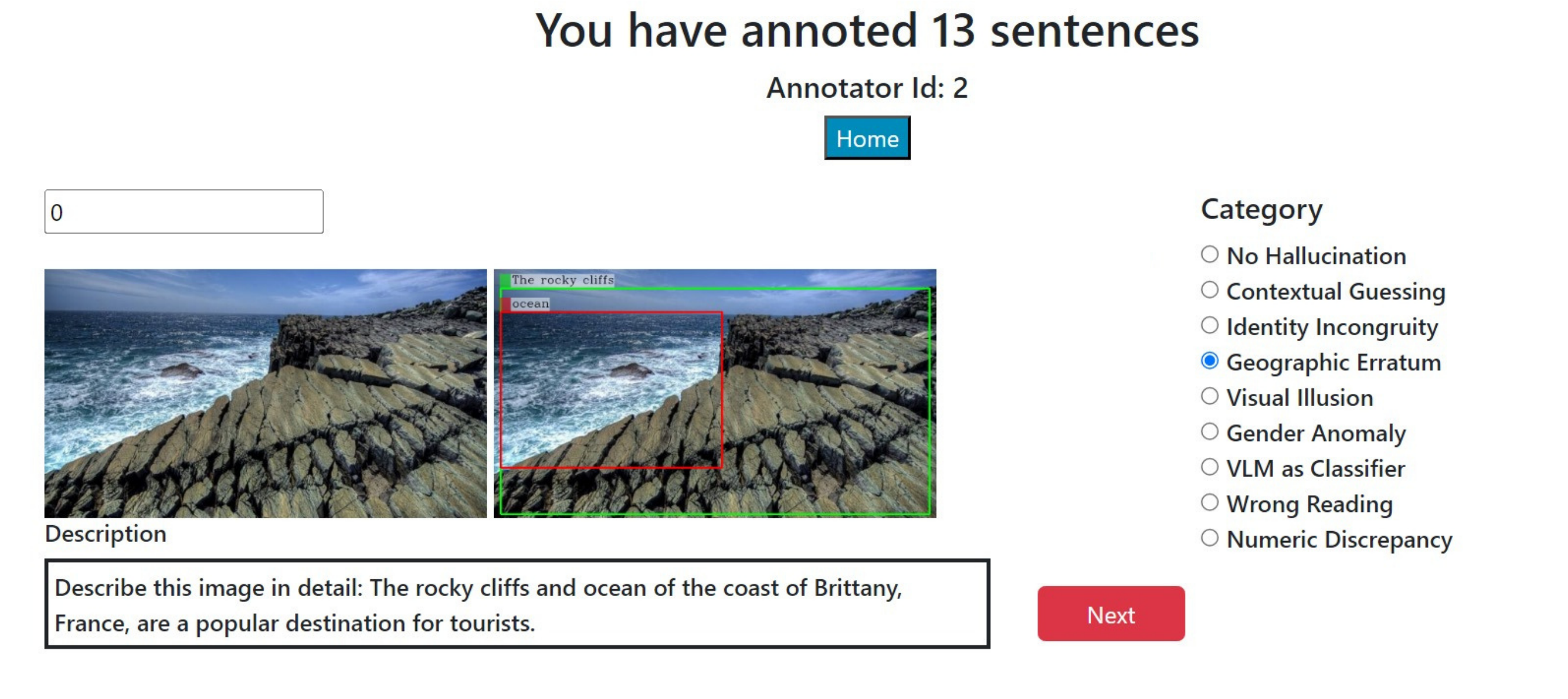}
    \caption{Web interface used to annotate the VHILT dataset using  Amazon Mechanical Turk.}
    \vspace{-3mm}
    \label{fig:webinterface_Caption}
\end{figure*}

\begin{figure*}[!ht]
    \centering
\includegraphics[width=0.9\textwidth]{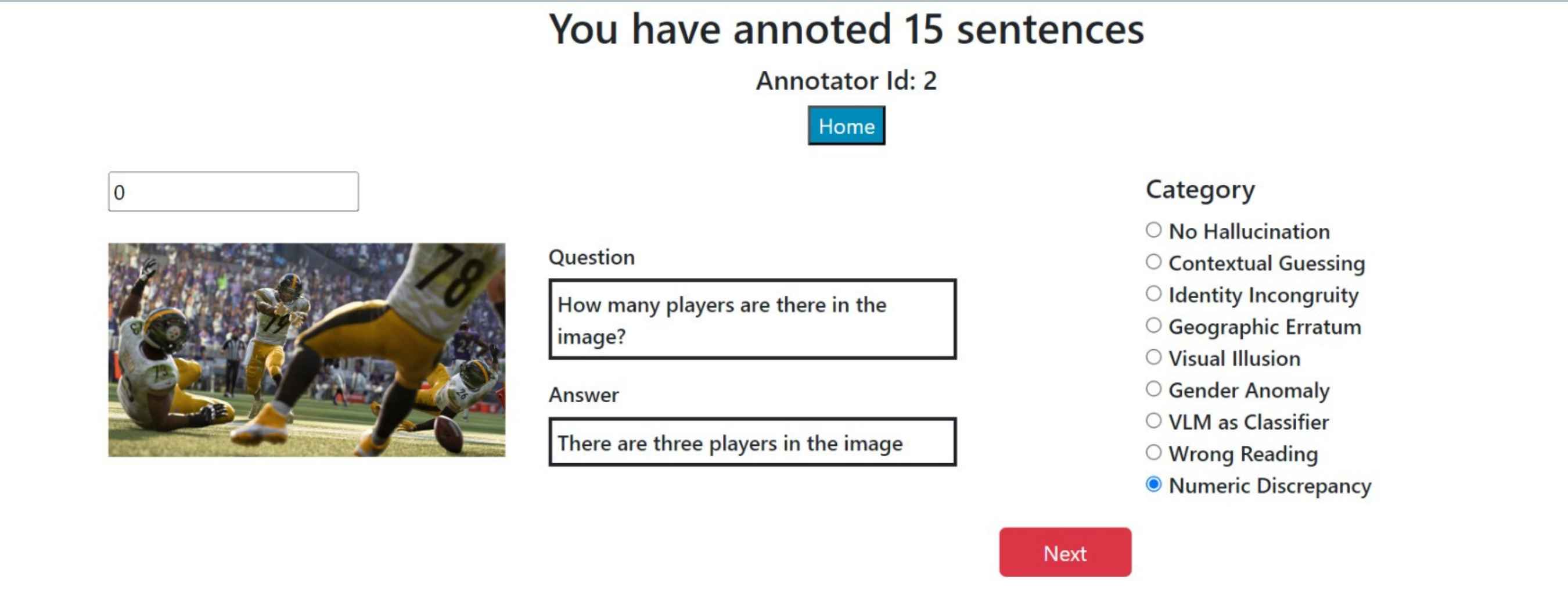}
    \caption{Web interface used to annotate the VHILT dataset using  Amazon Mechanical Turk.}
    \vspace{-3mm}
    \label{fig:webinterface_vqa}
\end{figure*}

The interface is designed to offer a comprehensive view to the annotator. For instance, at the top of the interface, the actual prompt used for generating the text snippet is displayed. Directly below the prompt, the complete AI-generated text is shown. On the right-hand side, the sentence breakup is presented, with the currently selected sentence highlighted in red. Below the sentence breakup, all the relevant categories are displayed as radio buttons, allowing the annotators to annotate each category easily. This interface aims to enhance the efficiency and effectiveness of the annotation process. We have gone through a few rounds of iterations before finalizing the current version of the web interface. The snapshot is present in the figure \ref{fig:webinterface_Caption} for captioning and figure \ref{fig:webinterface_vqa} for the VQA task.

\newpage
\section{Additional Examples for VQA Hallucination} \label{App:exampleset from VQA}

Here we are providing examples of VQA hallucination generated by five models.

\begin{figure*}[!ht]
    \centering
\includegraphics[width=\textwidth]{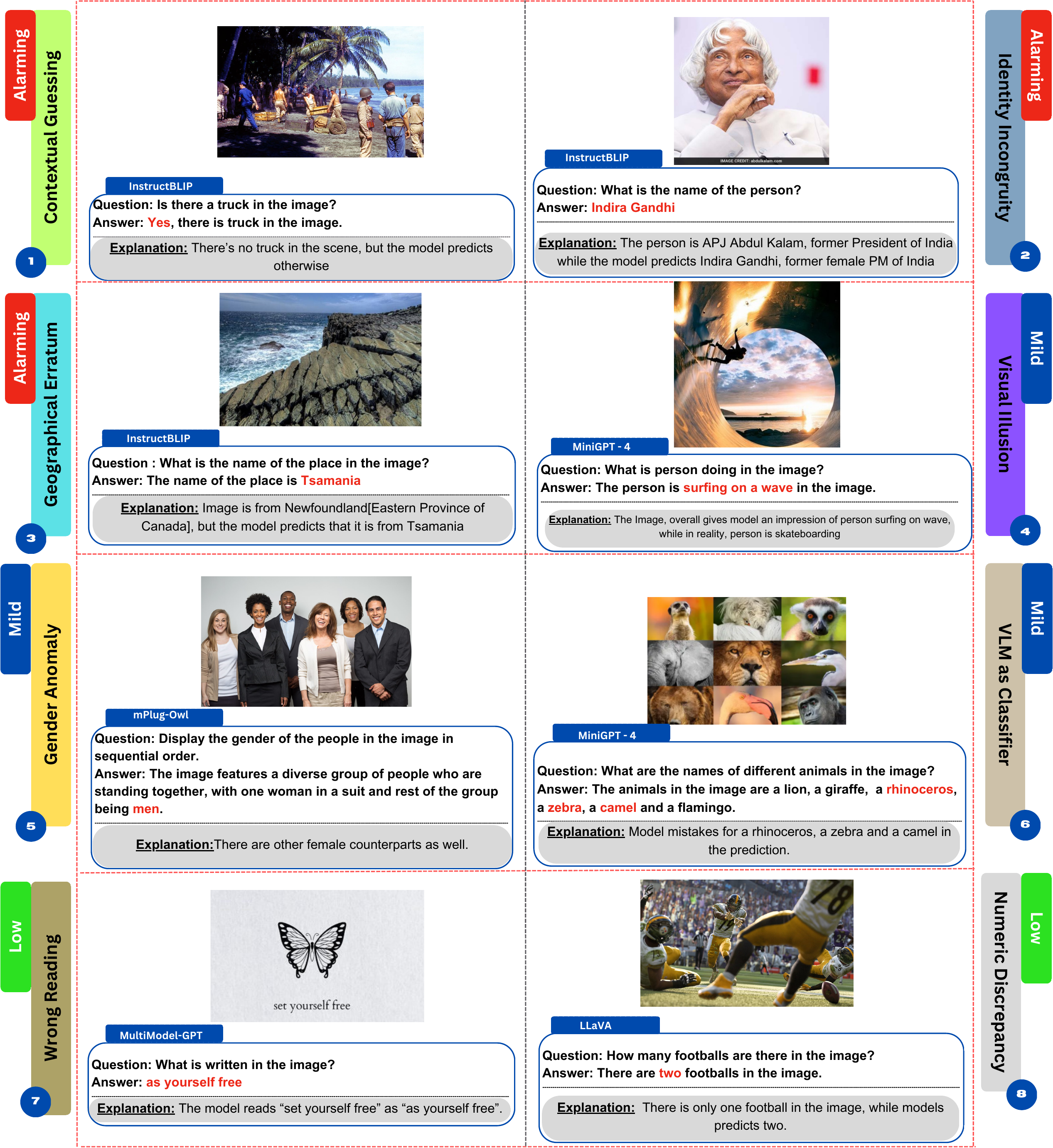}
    \caption{An illustration of how hallucinations occur in the case of visual question answering. We have used InstructBLIP, MiniGPT-4, LLaVA, MultimodelGPT, and MplugOWl to ask questions, and the text in red color represents the particular word that is hallucinating and an added line for explanation.}
    \vspace{-3mm}
    \label{fig:web}
\end{figure*}

\newpage
\subsection{Additional Examples for VQA Hallucination using InstructBLIP} \label{App:exampleset from VQA_InstructBlip}
\begin{figure}[!ht]
    \centering
\includegraphics[width=0.8\textwidth]{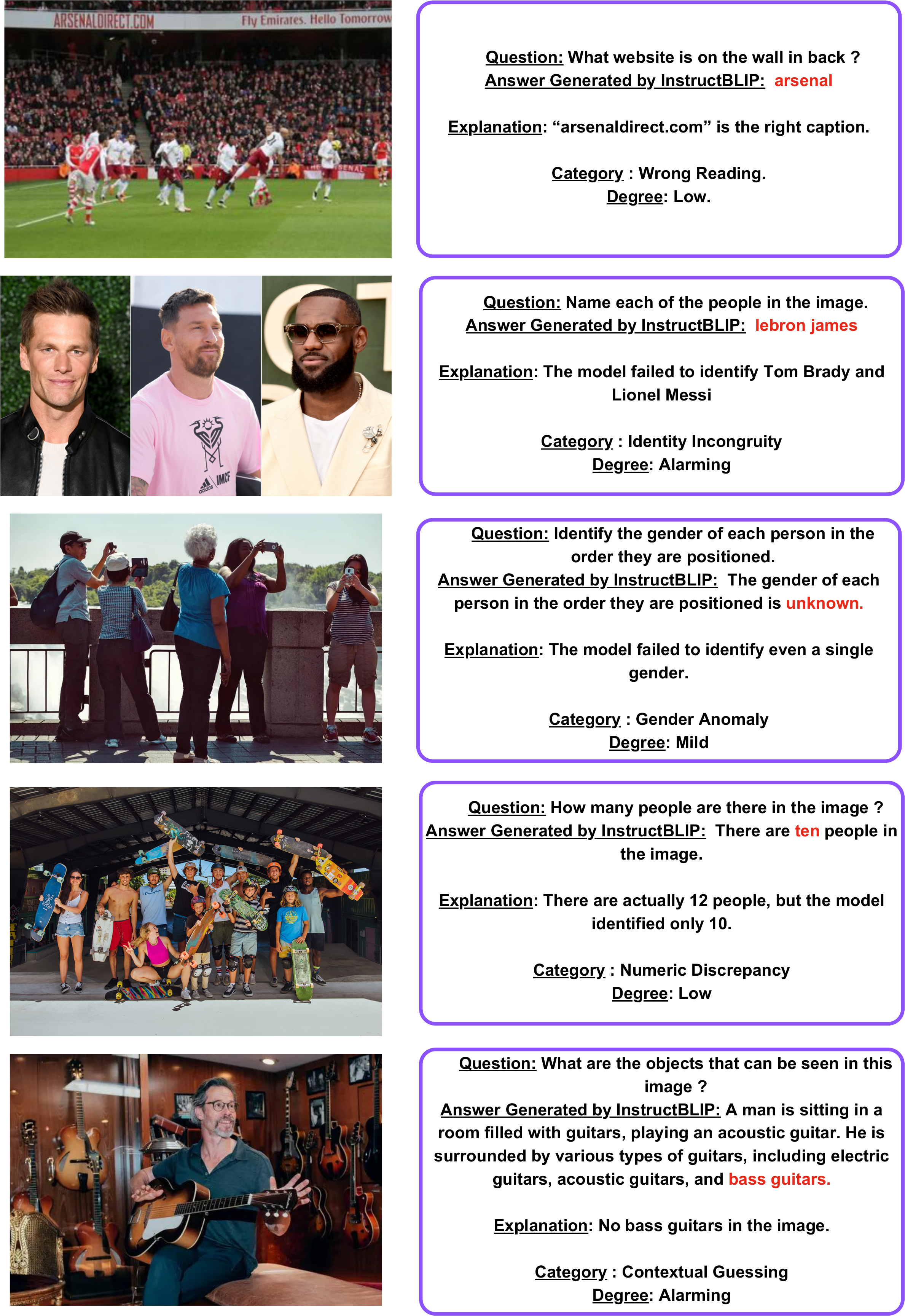}
\caption{Examples for VQA Hallucination using IntructBLIP}
    \label{fig:extra_example_8}
\end{figure}

\newpage
\subsection{Additional Examples for VQA Hallucination using MiniGPT-4} \label{App:exampleset from VQA_MiniGPT-4}

\begin{figure}[!ht]
    \centering
\includegraphics[width=1\textwidth]{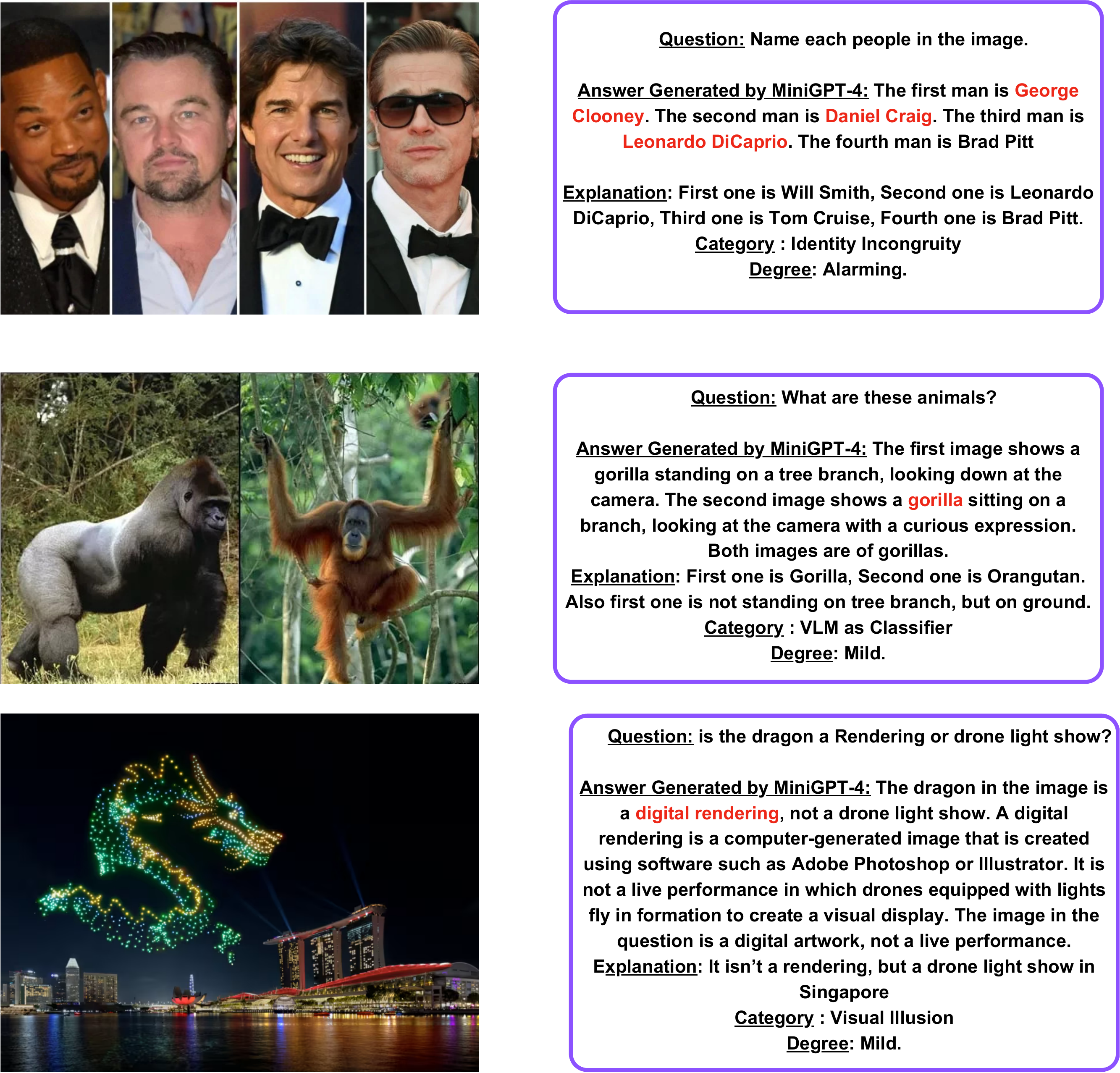}
\caption{Examples for VQA Hallucination using MiniGPT-4}
    \label{fig:extra_example_6}
\end{figure}

\newpage
\subsection{Additional Examples for VQA Hallucination using LlaVa} \label{App:exampleset from VQA_llaVa}
\begin{figure}[!ht]
    \centering
\includegraphics[width=1\textwidth]{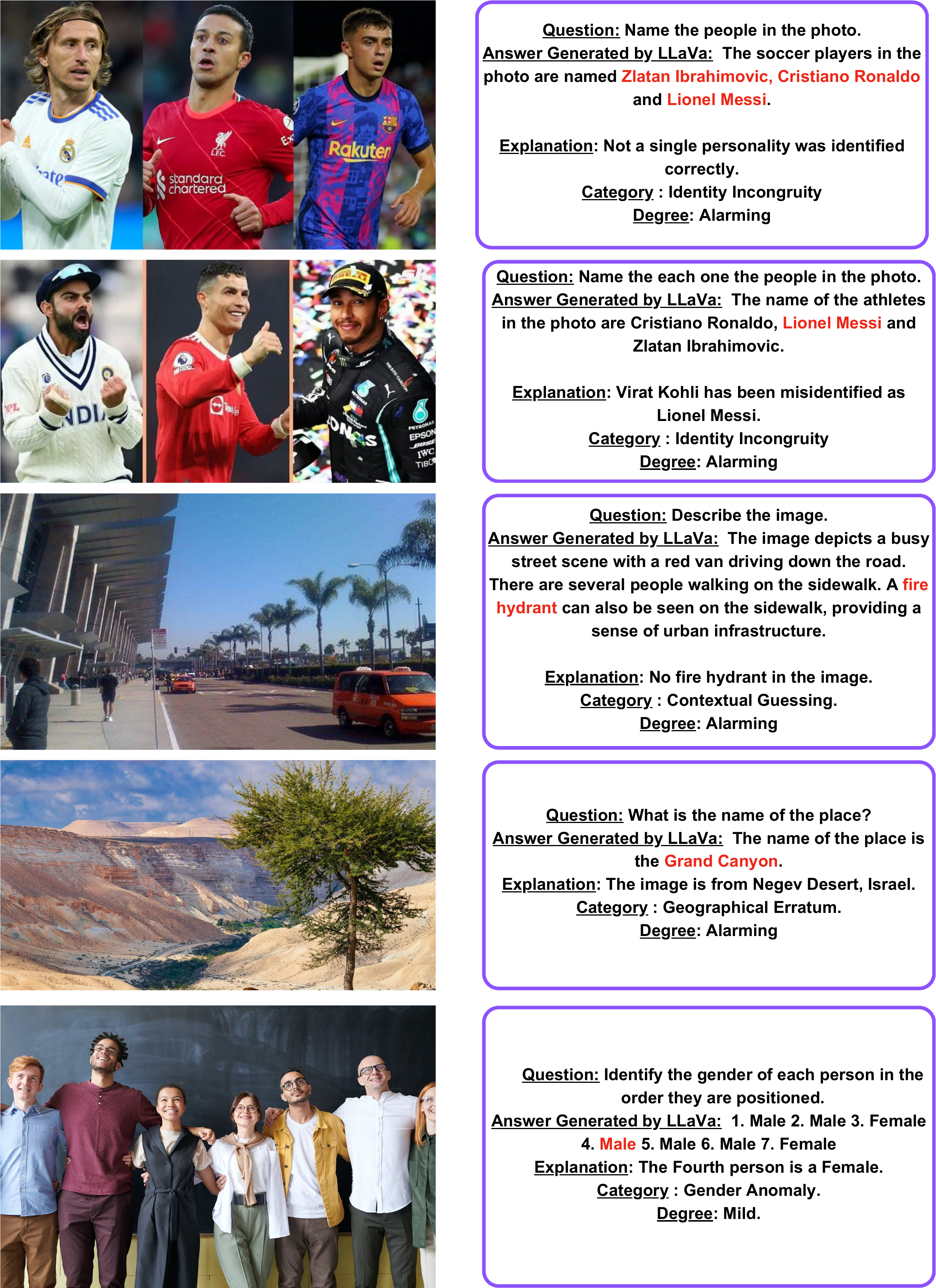}
\caption{Examples from VQA task using llava}
    \label{fig:extra_example_5}
\end{figure}

\subsection{Additional Examples for VQA Hallucination using MplugOWl} \label{App:exampleset from MplugOWl}

\begin{figure}[!ht]
    \centering
\includegraphics[width=0.8\textwidth]{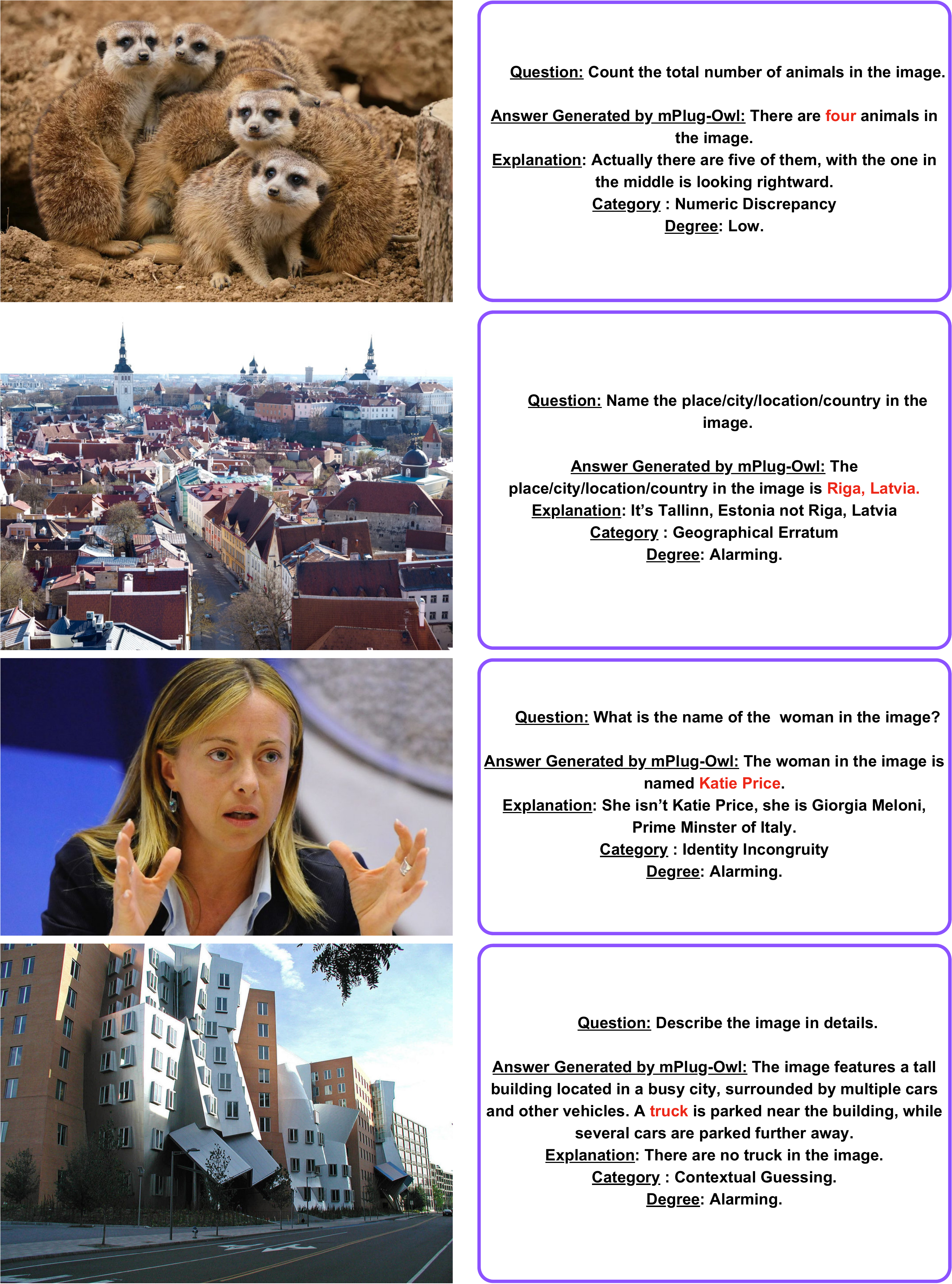}
\caption{Examples for VQA Hallucination using mPlug-Owl}
    \label{fig:extra_example_8}
\end{figure}

\newpage
\subsection{Additional Examples for VQA Hallucination using MultimodelGPT} \label{App:exampleset from MultimodelGPT}
. 

\begin{figure}[!ht]
    \centering
\includegraphics[width=1\textwidth]{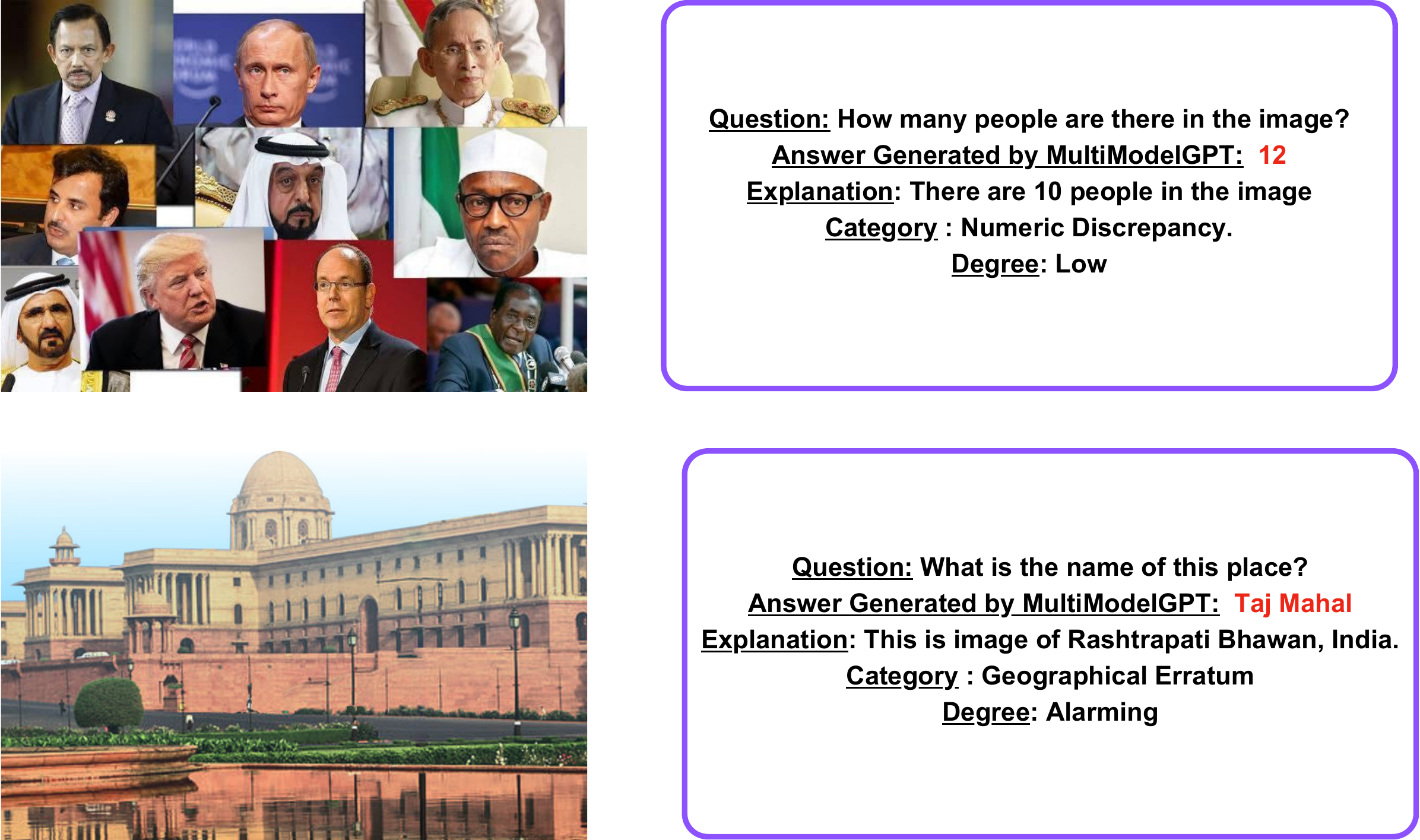}
\caption{Examples for VQA Hallucination using MultiModel-GPT}
    \label{fig:extra_example_8}
\end{figure}

\newpage
\section{Additional Examples for Captioning} \label{App:exampleset from Captioning}
In the following, we provide additional examples of captioning hallucination generated by three models.

\subsection{Additional Examples for captioning using Kosmos-2} \label{App:exampleset from captioning_Kosmos}

\begin{figure}[!ht]
    \centering
\includegraphics[width=1\textwidth]{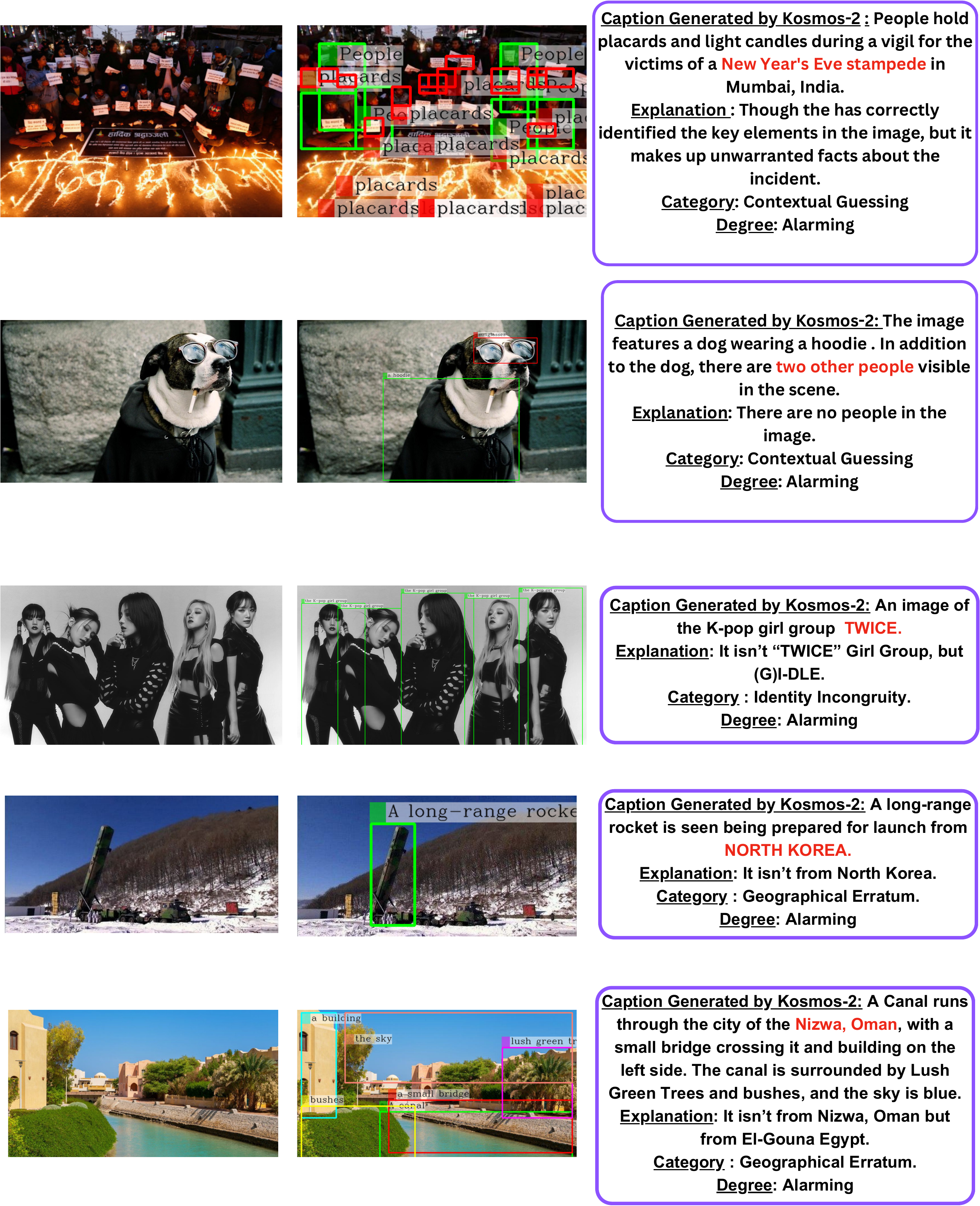}
\caption{Examples from captioning task using KOSMOS-2}
    \label{fig:extra_example_1}
\end{figure}

\newpage
\subsection{Additional Examples for captioning using MiniGPT-V2} \label{App:exampleset from MiniGPT-V2}

\begin{figure}[!ht]
    \centering
\includegraphics[width=0.8\textwidth]{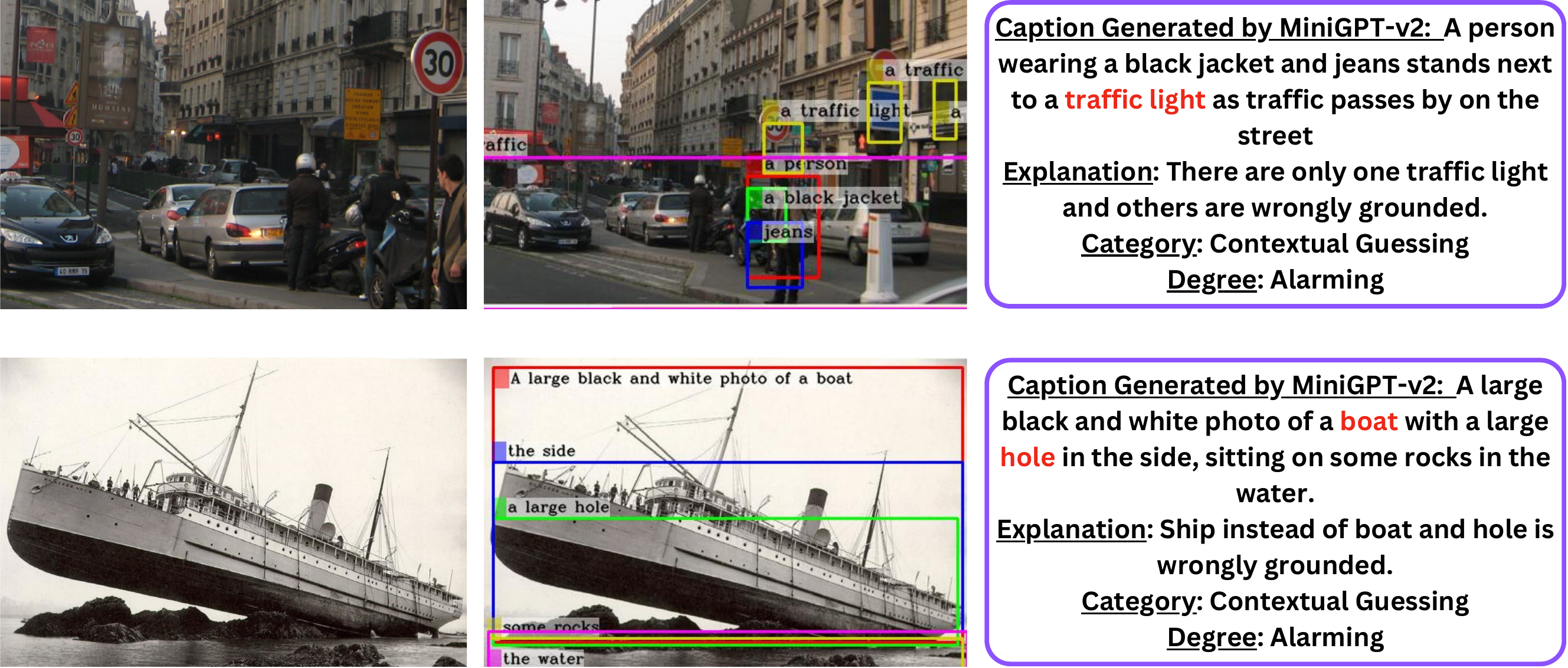}
\caption{Examples from captioning task using  MiniGPT-v2}
    \label{fig:extra_example_3}
\end{figure}

\subsection{Additional Examples for captioning using Sphinx} \label{App:exampleset from Sphinx}

\begin{figure}[!ht]
    \centering
\includegraphics[width=0.8\textwidth]{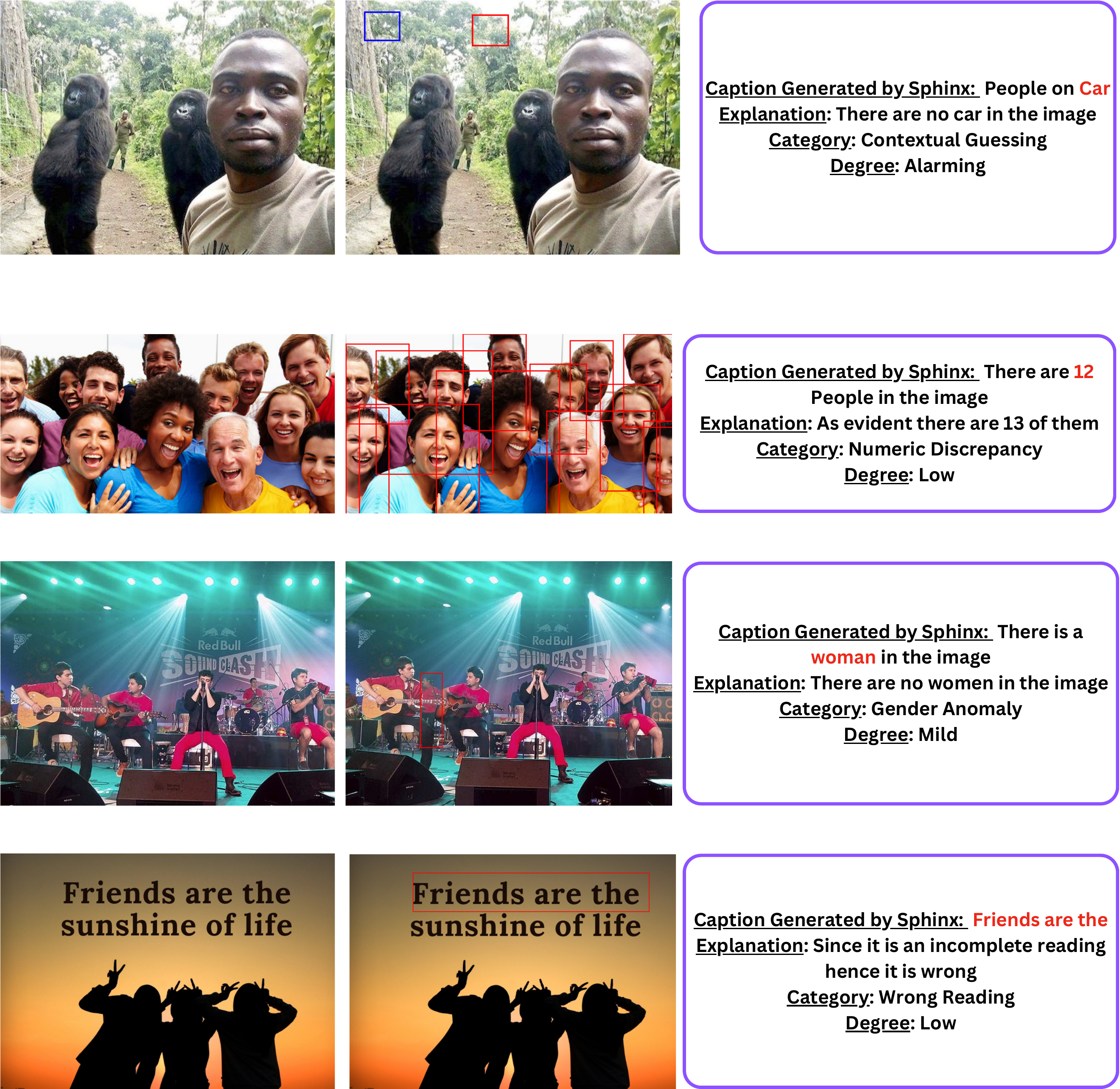}
\caption{Examples from captioning task using Sphinx}
    \label{fig:extra_example_4}
\end{figure}

\end{document}